\documentclass[letterpaper, 10 pt, conference]{ieeeconf}  %

\IEEEoverridecommandlockouts                              %

\usepackage{times}
\usepackage{epsfig}
\usepackage{graphicx}
\usepackage{amsmath}
\usepackage{amssymb}
\usepackage{booktabs}
\usepackage{pifont}%
\newcommand{\cmark}{\ding{51}}%
\newcommand{\xmark}{\ding{55}}%
\usepackage{blindtext}
\usepackage{multirow}
\usepackage[table,xcdraw,rgb]{xcolor}
\usepackage{tikz}

\usepackage[breaklinks=true,bookmarks=false]{hyperref}

\title{\LARGE \bf
ATTACH Dataset: Annotated Two-Handed Assembly Actions \\ for Human Action Understanding
}

\author{Dustin Aganian, Benedict Stephan, Markus Eisenbach, Corinna Stretz, and Horst-Michael Gross
\thanks{This work has received funding from the Carl-Zeiss-Stiftung as part of the project engineering for smart manufacturing (E4SM)}
\thanks{All authors are with Neuroinformatics and Cognitive Robotics Lab, TU Ilmenau, 98693 Ilmenau, Germany~~
{\tt\scriptsize dustin.aganian@tu-ilmenau.de}
}
}

\begin{document}

\maketitle
\thispagestyle{empty}
\pagestyle{empty}

\begin{abstract}
With the emergence of collaborative robots (cobots), human-robot collaboration in industrial manufacturing is coming into focus.
For a cobot to act autonomously and as an assistant, it must understand human actions during assembly.
To effectively train models for this task, a dataset containing suitable assembly actions in a realistic setting is crucial.
For this purpose, we present the ATTACH dataset, which contains 51.6 hours of assembly with 95.2k annotated fine-grained actions monitored by three cameras, which represent potential viewpoints of a cobot.
Since in an assembly context workers tend to perform different actions simultaneously with their two hands, we annotated the performed actions for each hand separately.
Therefore, in the ATTACH dataset, more than 68\% of annotations overlap with other annotations, which is many times more than in related datasets, typically featuring more simplistic assembly tasks.
For better generalization with respect to the background of the working area, we did not only record color and depth images, but also used the Azure Kinect body tracking SDK for estimating 3D skeletons of the worker.
To create a first baseline, we report the performance of state-of-the-art methods for action recognition as well as action detection on video and skeleton-sequence inputs.
The dataset is available at \href{https://www.tu-ilmenau.de/neurob/data-sets-code/attach-dataset}{https://www.tu-ilmenau.de/neurob/data-sets-code/attach-dataset}.
\end{abstract}

\section{Introduction}
\newcommand{\drawCam}[4]{
	\draw[thick, fill, rotate around={#3:(#1, #2)}, shift={(#1, #2)}] (-0.1, -0.15) -- (-0.1, 0.15) -- (0.1, 0.2) -- (0.1, -0.2) -- (-0.1, -0.15);
	\node at (#1,#2+0.4) {#4};
}

\newcommand{\imgwidth}{3cm}

\definecolor{workspacecolor}{rgb}{0.92, 0.95, 1}
\definecolor{workspaceborder}{rgb}{0.5, 0.55, 0.6}
\definecolor{backdrop}{rgb}{0.85, 0.85, 0.85}
\definecolor{shirtcolor}{rgb}{0, 0.35, 0.54}
\definecolor{haircolor}{rgb}{0.26, 0.20, 0.19}
\definecolor{tablecolor}{rgb}{0.65, 0.7, 0.75}

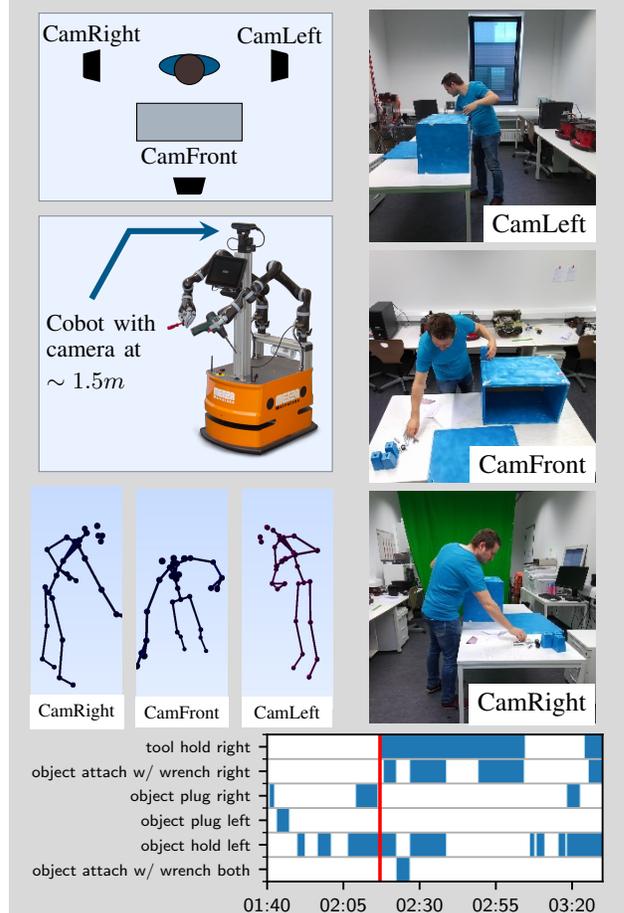
\begin{figure}[ht]
    \centering
    \begin{tikzpicture}[
        exarrow/.style={blue, line width=2mm},
        camdesc/.style={fill=white}
    ]
		\fill[backdrop] (-1.6, 3.2) rectangle (6.6, -9);
		
		\fill[workspacecolor] (-1.2, 3) rectangle (2.7, 0.5);
		\draw[workspaceborder] (-1.2, 3) rectangle (2.7, 0.5);
		
		\drawCam{-0.5}{2.3}{0}{\small CamRight};
		\drawCam{2}{2.3}{180}{\small CamLeft};
		\drawCam{0.8}{0.7}{90}{\small CamFront};
		
		\draw[fill=tablecolor] (0.1, 1.8) rectangle (1.5, 1.3);
		
		\draw[fill=shirtcolor] (0.8, 2.3) ellipse (0.4 and 0.15);
		\draw[fill=haircolor] (0.8, 2.25) circle (0.2);
		
		\node(left-example) at (4.7, 1.5){
		\includegraphics[width=\imgwidth]{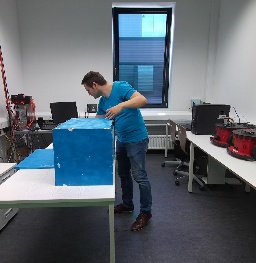}
		};
		\node[camdesc, anchor=east] at (6.2, 0.2){CamLeft};
		
		\node(front-example) at (4.7, -1.7){
		\includegraphics[width=\imgwidth]{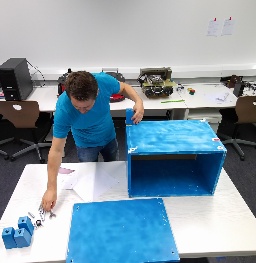}
		};
		\node[camdesc, anchor=east] at (6.2, -3){CamFront};
		
		\node(right-example) at (4.7, -4.9){
		\includegraphics[width=\imgwidth]{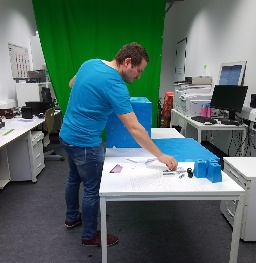}
		};
		\node[camdesc, anchor=east] at (6.2, -6.2){CamRight};
		
		\node(skel-front-example) at (0.7, -4.8){
		\includegraphics[width=1.2cm, trim={12cm 0 16cm 0}, clip]{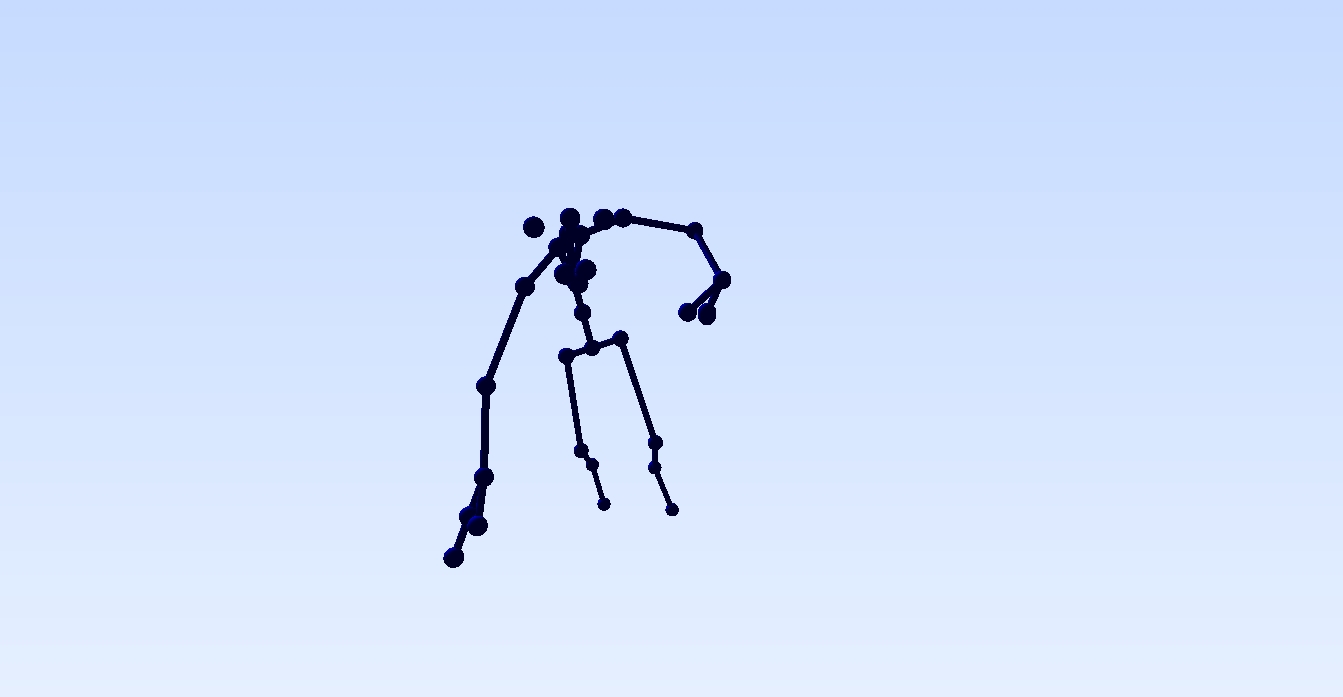}
		};
		\node[camdesc, minimum width=1.2cm] at (0.7, -6.3){\scriptsize CamFront};
		
		\node(skel-right-example) at (-0.7, -4.8){
		\includegraphics[width=1.2cm, trim={14cm 2.1cm 15cm 0}, clip]{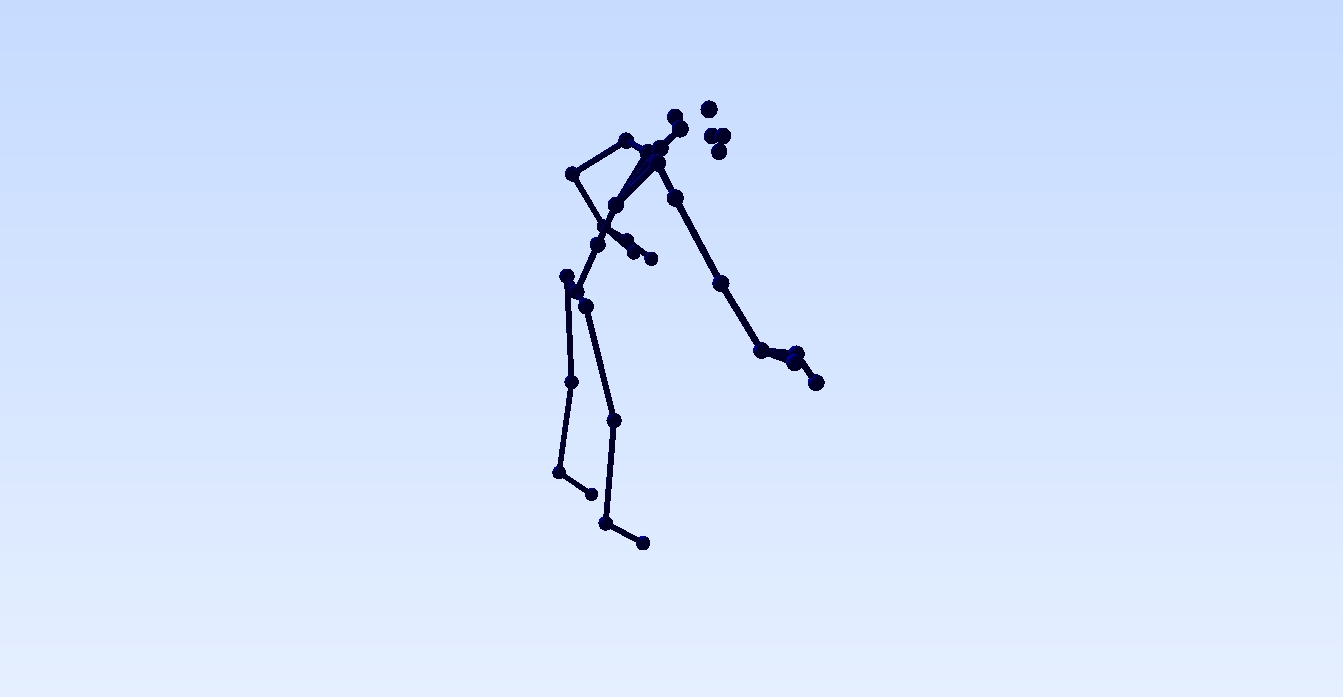}
		};
		\node[camdesc, minimum width=1.2cm] at (-0.7, -6.3){\scriptsize CamRight};
		
		\node(skel-left-example) at (2.1, -4.8){
		\includegraphics[width=1.2cm, trim={14cm 0 15cm 2.5cm}, clip]{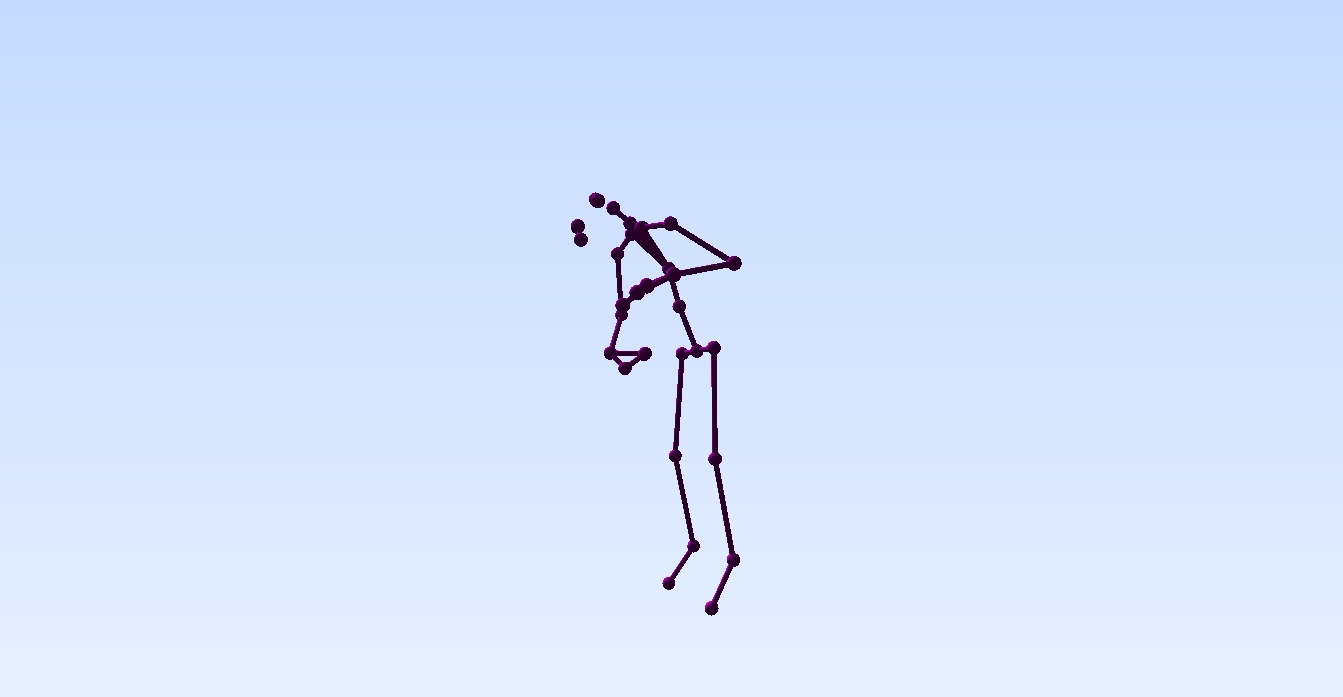}
		};
		\node[camdesc, minimum width=1.2cm] at (2.1, -6.3){\scriptsize CamLeft};
		
		\node at (2.5, -7.8){
		    \includegraphics[width=0.9\hsize]{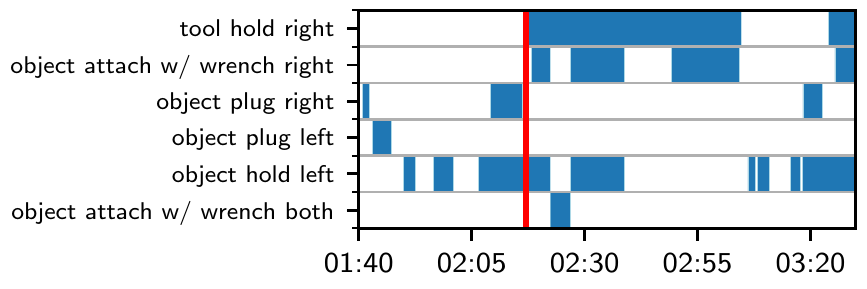}
		};

        \fill[workspacecolor] (-1.2, 0.3) rectangle (2.7, -3.1);
		\draw[workspaceborder] (-1.2, 0.3) rectangle (2.7, -3.1);
  
        \node at (1.5, -1.3){
            \includegraphics[width=0.27\hsize]{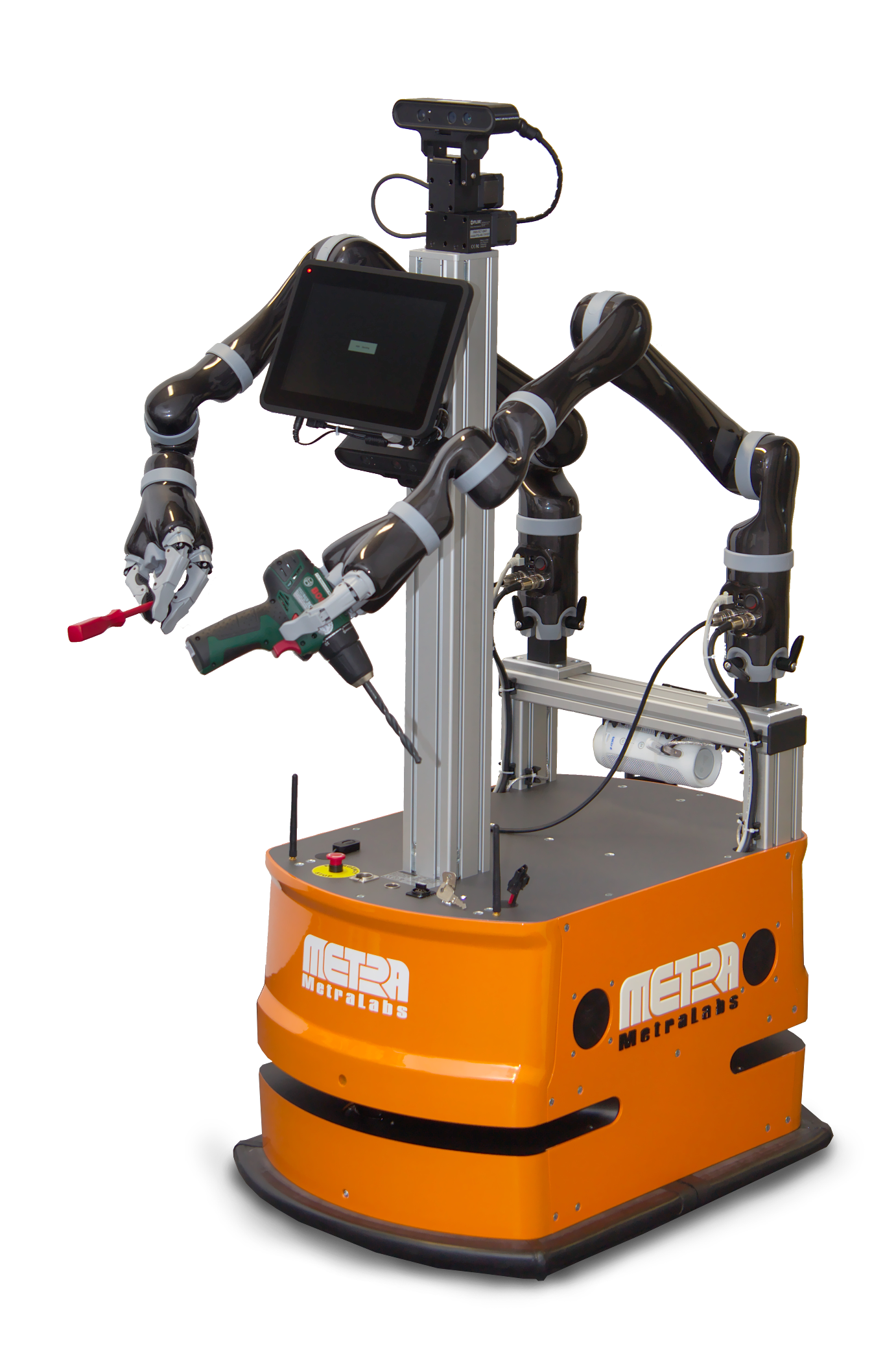}
        };

        \draw[-stealth, shirtcolor, line width= 0.8mm] (-0.5, -0.8) -- (0, 0.1) -- (1.2, 0.1);
        \node[text width=2cm, anchor=center] at (-0.1, -1.5) {\small Cobot with camera at \\ $\sim 1.5m$};
	\end{tikzpicture}
	\vspace{-4mm}
    \caption{Setup of the ATTACH dataset (top left), exemplary recorded data, and annotations (bottom).
    The views of the ATTACH dataset are representative of possible monitoring perspectives of our exemplary cobot.
    The red line in the annotation diagram (bottom) marks the timestamp at which the above images where recorded. The annotations show which actions were performed before (e.g., \textit{plugging in a leg}) and after (e.g., \textit{holding a wrench}) with each of both hands. 
    It can be seen that several actions temporally overlap, which is the focus of the proposed ATTACH dataset.
    }
    \label{fig:eye_catcher}%
    \vspace*{-6mm}%
\end{figure}
Versatile assembly in small quantities is of high relevance for small and medium-sized enterprises (SMEs).
For these, situation-aware cobots are currently a highly relevant research topic \cite{liu2022application,matheson2019human}.
As described in \cite{eisenbach2021}, one goal is to achieve situational awareness through general classification and detection algorithms for assembly, since it is not profitable to train new methods for each new manufacturing object.
In order to support the situation awareness of cobots, action recognition and action detection of the worker is a suitable tool.
However, most of the typical action recognition datasets published so far deal with daily activities~\cite{Dai2022TSU, Liu2020NTU, Smaira2020Kinetics}.
Action datasets that focus on manual activities usually deal with cooking actions~\cite{Damen2018EpicK, Damen2022EpicK100, Rohrbach2012MPIICooking, Zhou2018YouCook2} and only very recently with assembly actions~\cite{Shabat2021IKEA, Sener2022Assembly}, which are mostly single-label and, thus, only a single action is labeled at any given time for any small-grained action (e.g. \textit{pick up}, \textit{hold}, \textit{screw}).

However, to study situational awareness of assembly actions for cobots, we need a dataset where actions are performed in a natural way and thus can potentially occur simultaneously as the worker can perform an action with each hand.
Therefore, we present the novel ATTACH dataset (\underline{A}nno\underline{t}ated \underline{T}wo-Handed \underline{A}ssembly A\underline{c}tions for \underline{H}uman Action Understanding) for video- and skeleton-based action recognition and action detection during assembly.
For the ATTACH dataset, we asked 42 participants to assemble cabinets consisting of 26 parts, following three different sets of instructions. 
An overview of the assembly is given in Fig.~\ref{fig:eye_catcher}, which illustrates the three viewpoints we recorded.
Thus, we have 17.2 hours of recording time per view (51.6 hours in total) and an average duration of 8.2 minutes per recording (with 378 recordings in total).
Each fine-grained action (e.g., \textit{picking up a board with the left hand}, \textit{attach an object with a screwdriver in the right hand}) was annotated, resulting in a total of 95.2k annotations for 51 distinct action classes (with an average of 252 annotations per video).
During the recording of the ATTACH dataset, we focused on the following features:

\paragraph{Simultaneous fine-grained labels}
During assembly most workers often perform different actions simultaneously with their left and right hands, e.g., \textit{picking something up} with one hand and \textit{holding something} with the other hand (see Fig.~\ref{fig:eye_catcher}).
In contrast to previous single-label assembly action datasets, which do not represent such behavior, we did not restrict the participants to perform different actions with both hands and also labeled all available actions per hand for each time frame, as visualized in the lower part of Fig.~\ref{fig:eye_catcher}.
Thus, more than 54\% of all frames have more than one label describing the actions that occur and more than 68\% of annotations overlap with another annotation.

\paragraph{Diverse and dynamic assembly actions}
In creating the dataset, we took special care to ensure that participants received as few instructions as possible, i.e., they were not given a script to follow, as is often the case~\cite{Liu2020NTU, Sigurdsson2016Charades}.
Instead, they received only various written superficial instructions, such as those typically included with furniture for self-assembly.
Due to the variety of the parts to be assembled, actions also varied significantly in time.
They ranged from a fifth of a second for actions like \textit{lifting an object} or \textit{rotating a workpiece} to a minute or two for actions such as \textit{attaching an object with a wrench}.
Furthermore, each participant has a different level of craftsmanship, resulting in a very large variance in length and execution of the various actions.

\phantom{.}\\
We benchmark competitive methods on the ATTACH dataset on various tasks.
We evaluate action recognition of video clips and 3D-skeleton sequences, and focus on action detection on video and 3D-skeleton input.
We also evaluate the action detection task as a real-world robotic application. %

Furthermore, evaluating on skeleton input is necessary to evaluate the problem independently of the background.
This is important, since it is often not possible to directly record training data for the targeted environment and skeleton-based methods are mostly independent in this regard.
Therefore, we also apply an action detection method on skeleton sequences that previously has been applied on video data only.

Summarized, the main contributions of this paper are:
\begin{itemize}
    \item The publication of the ATTACH dataset, which is the first dataset to independently label each hand and thus include simultaneous fine-grained labels for the assembly action understanding problem.
    \item The application and evaluation of different baseline methods for the action recognition and action detection tasks for video-based and skeleton-based methods on the ATTACH dataset, to set a first baseline.
    \item Evaluation of an action detection method for an online robotic application on the ATTACH dataset.
\end{itemize}

\section{Related work}
\label{sec:related_work}
In the following, we present other related datasets for human action understanding.

\begin{table*}[ht]
\vspace{2mm}
\caption{Comparison of typical action recognition datasets and related action assembly datasets.}
\vspace{-2mm}
\resizebox{\textwidth}{!}{%
\scriptsize
\begin{tabular}{llllllllllll}
\hline
Dataset                                                      & \begin{tabular}[c]{@{}l@{}}Publ.\\ Year\end{tabular} & Activity                                                     & Frames                & Videos & \begin{tabular}[c]{@{}l@{}}Labelled\\ Instances\end{tabular} & Classes & \begin{tabular}[c]{@{}l@{}}Labelled\\ Frames\end{tabular} & \begin{tabular}[c]{@{}l@{}}Overlapping\\ Labels\end{tabular} & \begin{tabular}[c]{@{}l@{}}Partici-\\ pants\end{tabular}          & Views & Modalities                                                   \\ \hline
\rowcolor{lightgray!50!white!80} NTU RGB+D 120 \cite{Liu2020NTU}                                                      & 2019 & General                                                      & \multicolumn{1}{c}{-} & 114k & 114k  & 120     & \multicolumn{1}{c}{-}                                     & \xmark                                                       & 106                   & 3     & RGB, D, IR, 3DPose \\
Kinetics 700-2020 \cite{Smaira2020Kinetics}                                                    & 2020 & General YouTube                                                     & \multicolumn{1}{c}{-} & 455k  & 455k & 700     & \multicolumn{1}{c}{-}                                     & \xmark                                                       & - & 1     & RGB                                                          \\
\rowcolor{lightgray!50!white!80} Charades \cite{Sigurdsson2016Charades}                                                     & 2016 & Daily                                                        & \multicolumn{1}{c}{-} & 10k & 67k   & 157     & \multicolumn{1}{c}{-}                                     & \cmark                                                       & 267                   & 1     & RGB                                                          \\
TSU \cite{Dai2022TSU}                                                         & 2022 & Daily                                                        & 13.8M                 & 536  & 41k  & 51      & \multicolumn{1}{c}{-}                                     & \cmark                                                       & 18                    & 7     & RGB, D, 3DPose     \\
\rowcolor{lightgray!50!white!80}EPIC-KITCHENS-100 \cite{Damen2022EpicK100} & 2022 & Kitchen                                                      & 18M                   & 700 & 90k   & 4053    & 71.6\%                                                    & 28.1\%                                                       & 37                    & 1     & RGB                                                          \\
Meccano \cite{Ragusa2021Meccano}                                                     & 2021 & Toy Assembly       & 300K                  & 20   & 9k  & 61      & 84.9\%                                                    & 15.8\%                                                       & 20                    & 1     & RGB                                                          \\
\rowcolor{lightgray!50!white!80}Assembly101 \cite{Sener2022Assembly}                                                  & 2022 & Toy Assembly       & 111M                  & 4.3k & 1,014k  & 1380    & 81.4\%                                                    & 7.0\%                                                        & 53                    & 12    & RGB, 3DHandP       \\
IKEA ASM \cite{Shabat2021IKEA}                                                     & 2021 & Furniture Assembly & 3M                    & 371  & 17k  & 33      & 83.8\%                                                    & \xmark                                                       & 48                    & 3     & RGB, D, 3DPose     \\ \hline
\rowcolor{lightgray!50!white!80}ATTACH                                                       & 2023 & Furniture Assembly & 5.6M                  & 378  & 95k  & 51      & 91.3\%                                                    & 68.3\%                                                       & 42                    & 3     & RGB, D, IR, 3DPose \\ \hline
\end{tabular}
}
\vspace{-2mm}
\label{tab:datasets}
\end{table*}

\subsection{General and daily actions datasets}
Action recognition has become increasingly important in recent years, which is reflected in the large number of different datasets on various problems.
For instance, typical representatives for general actions as in NTU RGB+D are 60~\cite{Shahroudy2016NTU} and 120~\cite{Liu2020NTU}, which consist of video clips only a few seconds long, or the popular Kinetics dataset~\cite{Smaira2020Kinetics}, which is generated from YouTube videos and also deals with the classification of very short video clips.
On the other hand, many datasets are published in the field of household actions, such as DAHLIA~\cite{Vaquette2017DAHLIA}, Charades~\cite{Sigurdsson2016Charades}, Charades-Ego~\cite{Sigurdsson2018CharadesEgo}, Toyota Smarthome~\cite{Das2019ToyotaSmarthome} and TSU~\cite{Dai2022TSU}.
Here, the datasets differ strongly in their recorded perspectives (shooting, egocentric, monitoring) and the length variability of their actions (a few seconds in Charades versus a few seconds to several minutes in TSU).

\subsection{Cooking and instruction actions datasets}
In contrast to the assembly datasets, which have only recently been published, the datasets for instructions and for cooking \cite{Damen2018EpicK, Damen2022EpicK100, Miech2019Howto100m, Rohrbach2012MPIICooking, Tang2019Coin, Toyer2017IKEAFA, Zhou2018YouCook2} have been of research interest for an extended period of time.
However, these datasets are very strongly domain-related, and such domain-specific knowledge is not readily transferable.

\subsection{Assembly actions datasets}
So far, only very few papers have been published on assembly action datasets.
To the best of our knowledge, the only relevant ones are the toy assembly datasets Meccano~\cite{Ragusa2021Meccano} and Assembly101~\cite{Sener2022Assembly} and the furniture assembly dataset IKEA ASM~\cite{Shabat2021IKEA}, which is most related to our dataset.
Meccano as well as Assembly101 focus on the fine-motor assembly of toys.
Thus, the camera perspectives are focused on the hands and the assembly object.
In the case of Meccano, the assembly process was only recorded with an egocentric view.
Likewise, action understanding tasks using hand skeletons were a focus of Assembly101.
Their camera perspectives make these two datasets partially to not usable at all for cobot applications.

In contrast, we focus on the assembly task with a worker and a cobot who observes the worker and the workspace.
Therefore, an egocentric (worker-centric) perspective would not be suitable.
Additionally, the process of assembling a cabinet also requires finer movements, e.g., when \textit{screwing in the legs}, making it necessary to perceive both large body movements and smaller hand movements, when solving general action perception on our dataset.
When using cobots in such an assembly process, our camera setup also resembles the actual perspective of the robot more closely rather then the egocentric view of the worker.

The IKEA ASM dataset has a similar recording setup to ours.
The important difference, however, is that in the IKEA ASM dataset the natural behavior of the workers and thus simultaneously executed small-grained actions were not taken into account.
In contrast, we explicitly labeled with which hand which fine-grained action was performed resulting in simultaneous action labels.
Tab.~\ref{tab:datasets} summarizes the key points of our dataset compared to typical and relevant action datasets, clearly showing how we focused on the issue of simultaneous labels shown as the percentage of overlapping labels.

\section{Dataset}
In this section, we describe our overall setup for recording our dataset and give a basic overview of the statistics, such as length of annotated actions and dataset size.

\subsection{Setup}

The setup for data recording is shown in Fig.~\ref{fig:eye_catcher}.
A worktable is monitored by three Azure Kinect cameras\footnote{\scriptsize Technical specification: \href{https://docs.microsoft.com/en-us/azure/kinect-dk/hardware-specification}{https://docs.microsoft.com/en-us/azure/kinect-dk/hardware-specification}}, which capture the frontal, left, and right views of the worker assembling a piece of furniture, resembling typical observation positions of an assistive robot.
Each camera is connected to a separate PC and records RGB images with a resolution of $2560{\times}1440$ pixels and depth/IR images with a resolution of $320{\times}288$ pixels at $30\,$FPS.
Based on the known camera parameters, registered depth images with the same resolution as the RGB images can be calculated.
Extrinsic calibration is realized by a cube with ArUco markers~\cite{garrido2014automatic} placed in the center of the worktable before a recording took place.
This position marks the center of the global coordinate system.
For all images captured, we recorded their globally synced timestamps, enabling to match corresponding images across views if necessary.
For each camera, the Azure Kinect body tracking SDK is employed, which uses the depth and IR images to extract a 3D skeleton of the worker.

\subsection{Data and Annotations}

\paragraph{Assembly task}
In each recording, the furniture to be assembled are IKEA cabinets, each consisting of 26 parts.
We created three versions of the assembly instructions, which differed in the order of the assembly steps and the actions to be performed, such as performing certain actions with bare hands or with a tool.
Each of the recorded subjects had to assemble the piece of furniture according to the construction manual.
As our manuals only consisted of goal-oriented instructions, like in furniture assembly manuals, we did not specify how to achieve the next step.
On average, it took the participants 8.2 minutes to assemble the piece of furniture.
Overall, we recorded three complete assemblies for all 42 participants from three different viewpoints each resulting in 378 recordings and 51.6 hours of recordings in total.

\paragraph{Participant statistics}
We recorded 42 participants with different level of experience in assembling of which 31 were male and 11 were female.
The age of the participants ranged from 21 to 67.

\begin{figure}[t]
    \centering
    \includegraphics[width=0.98\linewidth]{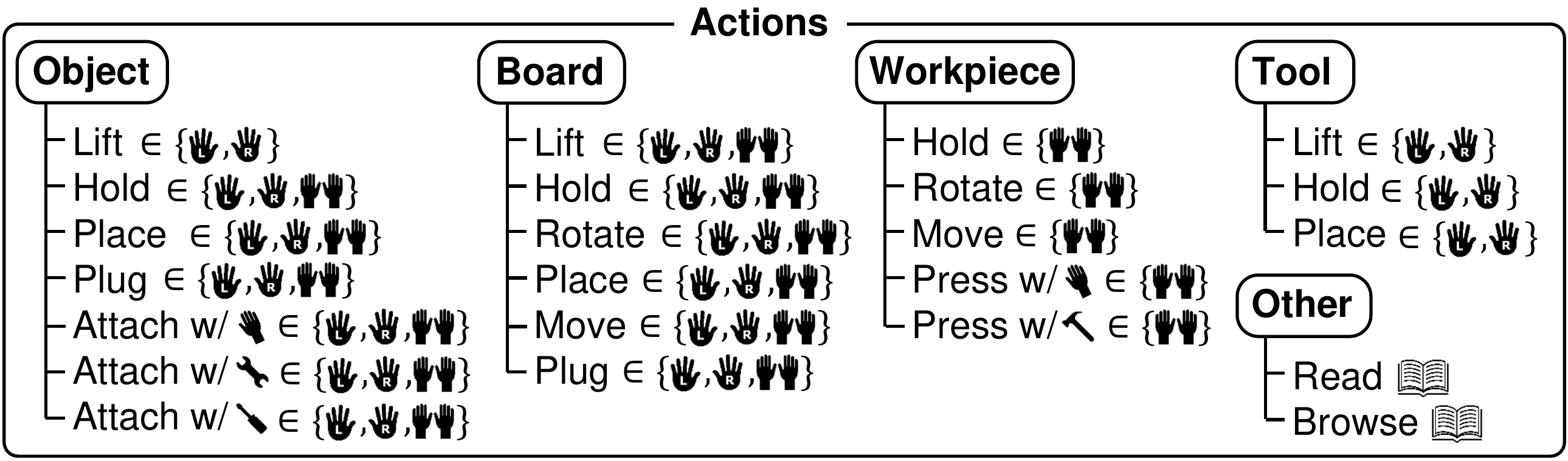}
    \vspace{-2mm}
    \caption{Action classes used for annotations.}
    \label{fig:actionLabels}
    \vspace{-4mm}
\end{figure}

\paragraph{Annotations}
The recorded data is annotated in detail, as shown in Fig.~\ref{fig:actionLabels}.
The type of object on which a particular action is performed is distinguished.
This is necessary for follow-up tasks to identify specific assembly steps.
We distinguish actions performed on five types of objects:
"Object" in Fig.~\ref{fig:actionLabels} is a small object, such as a screw, that can be enclosed by one hand and of which it is easy to hold multiple instances in one hand.
Actions performed on the walls of the cabinet or the like are included in the "board" category.
Actions performed on partially assembled furniture are grouped in the "workpiece" category.
When the subject uses a tool, the corresponding action is placed in the respective category, unless it is applied directly to a specific object or workpiece.
In that case, we annotated it as attaching an object with a specific tool or pressing with a tool.
The category "other" contains actions that are performed with the construction manual (e.g. \textit{reading}, \textit{browsing}).

\begin{figure}
    \centering
    \vspace{2mm}
    \includegraphics[width=0.95\hsize]{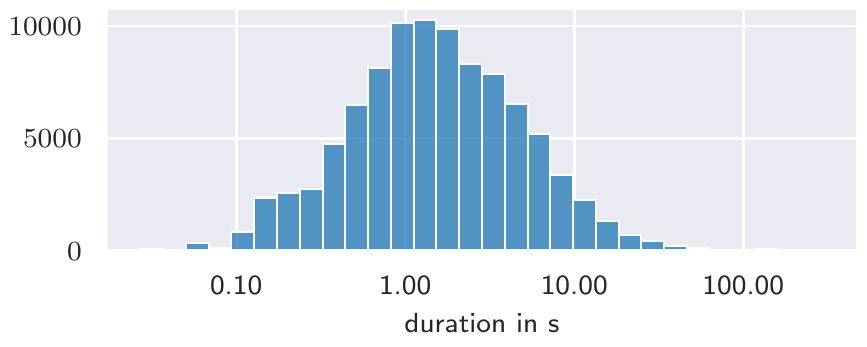}
    \vspace{-4mm}
    \caption{Histogram of action duration in our dataset plotted on a logarithmic scale. The longest action took 299 seconds ($\sim5$ min).}
    \label{fig:action_duration_hist}
    \vspace{-4mm}
\end{figure}

Using this scheme, we get 51 action classes as shown in Fig.~\ref{fig:actionLabels}.
For each category, we annotated several actions separately for both hands.
This means, that the subject can perform one action with the left hand while simultaneously performing another action with the right hand, e.g., as shown in Fig.~\ref{fig:eye_catcher}.
This results in more than one label for 54\% of all frames and more than 68\% of annotations overlapping with another annotation.
Overall, the data are annotated with 95.2k annotations, which corresponds to 252 annotations per assembly sequence on average.
A histogram of the duration of the performed actions in our dataset can be found in Fig.~\ref{fig:action_duration_hist}.

\subsection{Dataset splits}

For evaluations on our dataset, we use a person and a view split, similar to other datasets like TSU~\cite{Dai2022TSU}.

\textit{Person-split}: We split our participants into three groups, with recordings from two-thirds of all participants (28) used for training and the remaining third split into validation (4 participants) and test data (10 participants).
Care was taken to ensure that all action classes appear with sufficient frequency in both the test and training splits.

\textit{View-split}: The camera views shown in Fig.~\ref{fig:eye_catcher} were split as follows:
\textit{CamRight} was used as test data, while recordings from \textit{CamFront} and \textit{CamLeft} were used for the training and validation splits.
As the views already have a drastically different perception of the scene, we chose not to assign another separate camera for validation.
Instead, 10\% of all recordings from the front and left camera were assigned for validation.
As we will show in our experiments below, splitting the view is a major challenge because the scene looks vastly different from each point of view.
Furthermore, in at least one of the views, the person and the action performed are always partially obscured by furniture parts.
This split represents the situation of a mobile cobot viewing the scene from a different perspective than those available during training.
\section{Experiments on action recognition}
\label{sec:action_recognition}

\begin{figure*}
    \centering
    \vspace{2mm}
    \includegraphics[width=\hsize]{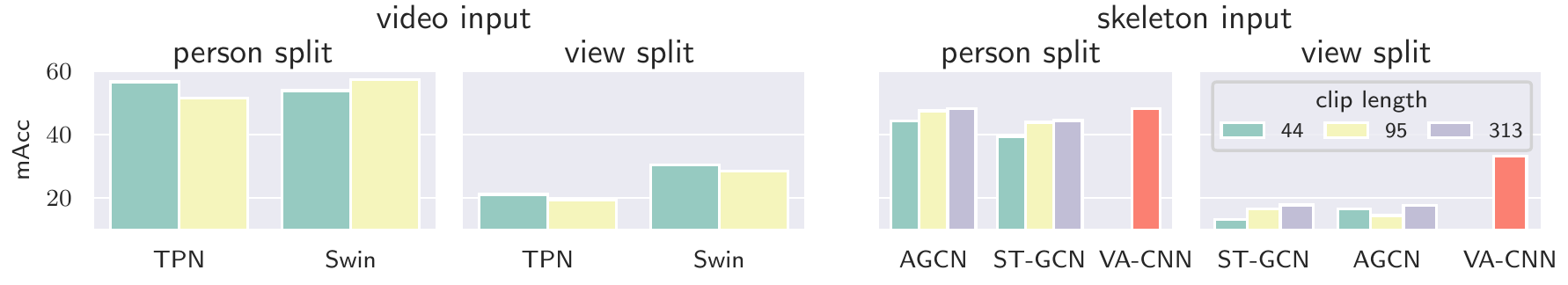}
    \vspace{-7mm}
    \caption{Baseline results on the action recognition task for different modalities and methods. We only report the mean class accuracy. The top-k metrics for the best models are reported in Tab.~\ref{tab:recognition_results}.
    VA-CNN does not require a fixed number of input frames as the trimmed skeleton-sequence is scaled to size.}
    \label{fig:recognition_results}
    \vspace{-4mm}
\end{figure*}

In the following, we benchmark state-of-the-art methods on our dataset on the typical action recognition task.
Although action detection is more suitable for online robotic applications, this evaluation is still needed because action detectors typically use action recognition methods as a backbone for feature estimation.
Thus, we provide first baselines of state-of-the-art methods chosen based on their model architectures (e.g. 3D-CNN vs. transformer) and for our two modalities (video vs. skeleton sequences) respectively.

\subsection{Evaluation protocol}
We report the typically used metrics mean class accuracy (mAcc), the top1, and the top5 accuracy on trimmed video clips and trimmed skeleton sequences.
We use this evaluation to determine which of the trained networks to use as the backbone for the action detection task (Sec.~\ref{sec:action_detection}).

\subsection{Video-based approaches}
\paragraph{Setup}

As state-of-the-art methods for action recognition on video-sequence inputs, we decided to use TPN~\cite{Yang2020TPN}, a well performing 3D-CNN-based approach, and the novel swin video transformer (Swin)~\cite{Liu2022Swin}.
Implementations for both methods are publicly available\footnote{\scriptsize \href{https://github.com/SwinTransformer/Video-Swin-Transformer}{https://github.com/SwinTransformer/Video-Swin-Transformer}} and are based on MMAction2~\cite{2020mmaction2}.

For the hyperparameter search during training, we used the original values of the respective methods as a starting point.
Furthermore, as the actions performed in our dataset differ in length, compared to the datasets the methods were originally trained with a clip length of $16$, we also tested the mean action length of $95$ and the median action length of $44$ frames respectively.
For the process of creating clips from segments for training or evaluation, we have strictly adhered to the original implementations and the usual state-of-the-art practices.

\paragraph{Results}
Fig.~\ref{fig:recognition_results} shows the results of our baselines on video input.
A significant difference between the person and view split is observable.
While Swin and TPN perform comparably on the person split, for the view split Swin outperforms TPN which suggests Swin to be able to generalize better across different views.
Generalization across people seems to be the easier task as the performance of both methods is around twice as high as on the view split.

For the different clip lengths tested, Swin performed best with the mean clip length of $95$ for the person split.
Otherwise, the median clip length of $44$ was slightly better.

In direct comparison, Swin outperforms TPN (by a large margin on the view split) on our dataset, making it the better choice as a feature extractor for the action detection methods (evaluated in Sec.~\ref{sec:action_detection}).

\subsection{Skeleton-based approaches}
\paragraph{Setup}
For training skeleton-based approaches, we use the 3D skeleton keypoints estimated by the Azure Kinect body tracking SDK which consists of $32$ joints.
A visualization of these skeletons can be found in Fig.~\ref{fig:eye_catcher}.

As state-of-the-art methods for action recognition on skeleton-sequence inputs, we decided in favor of VA-CNN~\cite{Zhang2019Vacnn} as a 2D-convolution-based (2D-CNN) approach, and ST-GCN~\cite{Yan2018STGCN} as well as AGCN~\cite{Shi2019AGCN} as graph-convolution-based (GCN) methods.
We used publicly available code for VA-CNN\footnote{\scriptsize \href{https://github.com/microsoft/View-Adaptive-Neural-Networks-for-Skeleton-based-Human-Action-Recognition}{https://github.com/microsoft/\\View-Adaptive-Neural-Networks-for-Skeleton-based-Human-Action-Recognition}} and MMaction2~\cite{2020mmaction2} for ST-GCN and AGCN.

For the hyperparameter search during training, again, we oriented on the parameterization in the original implementations as a starting point.
Likewise, the skeletons were normalized according to each of the applied methods.

For VA-CNN, the complete trimmed skeleton sequences were transformed into a three channel image (x, y, z) and resized to a resolution of $224{\times}224$ pixels ($\text{keypoints}{\times}\text{frames}$).
Thus, no fixed clip length had to be used for training.

For the GCN methods, we need to train with fixed clip lengths, so similarly to the video-based methods, we chose $44$ (median action length) and $95$ (mean action length).
Furthermore, we also train with the clip length of $313$, which corresponds to the 95\% quantile of action lengths.
In addition, $313$ is close to the clip length of $300$, which is often used for GCNs.

\paragraph{Results}
The results of our skeleton-based baselines are shown in Fig.~\ref{fig:recognition_results}.
Roughly speaking, an increase in clip length results in an increase in recognition performance. 

For the person split, VA-CNN and AGCN perform similarly well, with ST-GCN being slightly worse.
In contrast, the view split shows a different trend.
Here, VA-CNN is clearly superior to GCN methods.
This is probably due to the view-adaptive module of VA-CNN, which is supposed to learn a normalization of the skeletal view.
However, we also trained the VA-CNN without the view adaptive module on the view split and achieved a mAcc of 22.2\%, which is still more than four percentage points better than the best GCN results.
This shows, that the examined GCN methods have more difficulties in generalizing a view than the applied CNN method.
Thus, VA-CNN is the better choice as a feature extractor for action detection methods on skeletons (evaluated in Sec.~\ref{sec:action_detection}).

When comparing the results of the video-based and the skeleton-based methods, on the person split -- as expected -- the former perform better.
However, the view-split comparison also demonstrates that VA-CNN generalizes even better across views than Swin.
This indicates that skeleton-based methods are capable of good generalization.
This further highlights the difficulty of the view split and the significant visual differences between the training and test data.

\begin{table}
    \centering
        \caption{
    Summary of results for action recognition for video- and skeleton-sequence-based inputs.
    The clip length indicates the best clip length for the respective model and split.
    mAcc represents mean class accuracy and top1 and top5 represent top-k accuracy.
    }
    \vspace{-2mm}
    \scriptsize
        \begin{tabular}{lllrrrr}
\toprule
         &       &      &  clip length &  mAcc &    top1 &    top5 \\
 & model & split &           &                 &         &         \\
\midrule
\rowcolor{lightgray!50!white!80} video &  Swin & person &        95 &            \textbf{57.4} &  \textbf{61.3} &  \textbf{92.1} \\
\rowcolor{lightgray!50!white!80}         &        & view &        44 &            \textbf{30.4} &  \textbf{33.8} &  \textbf{75.1} \\
        & TPN & person &        44 &            56.6 &  59.3 &  90.7 \\
         &        & view &        44 &            21.2 &  27.0 &  66.5 \\ \midrule
\rowcolor{lightgray!50!white!80} skeleton & AGCN & person &       313 &            48.2 &  55.7 &  86.9 \\
\rowcolor{lightgray!50!white!80}         &         & view &       313 &            17.7 &  22.0 &  55.5 \\
         & ST-GCN & person &       313 &            44.5 &  51.6 &  84.8 \\
         &        & view &       313 &            17.8 &  22.6 &  58.5 \\
\rowcolor{lightgray!50!white!80}        & VA-CNN & person &        - &            \textbf{48.2} &  \textbf{56.5} &  \textbf{87.2} \\
\rowcolor{lightgray!50!white!80}         &        & view &        - &            \textbf{33.2} &  \textbf{43.0} &  \textbf{76.9} \\\bottomrule
\end{tabular}
    \label{tab:recognition_results}
\vspace{-6mm}
\end{table}

\section{Experiments on action detection}
\label{sec:action_detection}
In action detection, for each frame every occurring class has to be detected.
In the following, we begin by presenting our robotic application scenario for action detection.
We then briefly describe the chosen method, the evaluation protocol, and the training setup, followed by our experiments for offline and online action detection to get a first baseline on the presented ATTACH dataset.

\subsection{Robotic application scenario}
As described in \cite{eisenbach2021}, our goal is to use autonomous cobots to assist workers during assembly.
In order to achieve this goal, cobots must be able to recognize what is happening live and react to it.
Using the ATTACH dataset, we aim to train and evaluate the cobots' environmental awareness regarding action detection.
We will first present the experimental results for the typical offline usage, where the complete recording is available during the detection.
During offline usage, the detection for a frame is often based not only on past and present frames but also on future frames.
Since a cobot is supposed to detect actions live and not after a few minutes, such a detection is not practical for our use case.
Therefore, we also do an online evaluation, where only the present and past frames are available for the detection task on each frame.

\subsection{Action detection method}
For the action detection task we decided to employ the Pyramid Dilated Attention Network (PDAN)~\cite{Dai2021PDAN}~\footnote{\scriptsize Publicly available code: \href{https://github.com/dairui01/PDAN}{https://github.com/dairui01/PDAN}} with the chosen recognition methods from experiments in Sec.~\ref{sec:action_recognition} as feature extractors.
Unlike many other methods, PDAN is well suited to capture the temporal relationships of both short and long simultaneous actions.
This is shown by the state-of-the-art results on the TSU dataset~\cite{Dai2022TSU}, which has simultaneous fine-grained labels for daily activities. %

PDAN processes non-overlapping clips of frames of fixed size.
Based on these clips, an encoder is used to generate a feature vector for every clip.
PDAN then predicts for every frame the occurring actions.
In the following experiments, we tested different clip lengths with the minimum clip length being $8$ frames.
Taking the limitation of our embedded hardware (Jetson Xavier) into account, this minimum length gives enough time to complete the detection on the last clip, before the next clip has to be processed.

PDAN, like many other similar action detectors, has only been used on video-based feature extractors.
As shown in our previous experiments, skeleton-based methods tend to generalize better over different views.
This is advantageous to our robotic scenario as a cobot can be mobile and therefore has a non-fixed perspective.
Thus, we also apply PDAN on the skeleton-based features.

\subsection{Evaluation protocol}
Dai \textit{et al.} \cite{Dai2021PDAN} evaluated PDAN on the TSU dataset~\cite{Dai2022TSU} whose varying action duration and overlapping labels make it comparable to our dataset.
Therefore, we evaluate the action detection task using the same frame-based mAP evaluation protocol as used in \cite{Dai2021PDAN}.
This performance measure resembles a frame-based accuracy averaged over all action classes, making it suitable for our unbalanced dataset.

\subsection{Setup}
As PDAN cannot be applied to video inputs directly, but needs a feature representation generated by an encoder, we first have to choose the models for this purpose.
For comparison to the original implementation, we report the detection results when employing the originally used feature extractor I3D~\cite{Carreira2017I3D} as described in \cite{Dai2021PDAN}.
This encoder is trained on clips consisting of $16$ frames from the Charades dataset.
In addition, we use the best models for both modalities described in Sec.~\ref{sec:action_recognition}, namely the video swin transformer (Swin) and VA-CNN on the respective splits.
We like to highlight that this is the first attempt to use PDAN with features that were not extracted from video, but instead from skeleton sequences.
Swin and VA-CNN were not only trained on clips consisting of $16$ frames and are also directly optimized on our dataset which should enable an improvement in detection quality.

During the hyperparameter search for training, in addition to the typical parameters and the clip size, we also optimized the PDAN-specific model parameters for the temporal reception field.

\subsection{Results for offline usage}
The results of our trained action detection models for offline usage are shown in Fig.~\ref{fig:results_detection}.
As can be seen, the difference across both splits is similar to the models trained on the action recognition task, with the view split being more difficult than the person split.
An interesting fact is the impact of the clip length on the detection performance.
A significant difference can be observed between the video-based and the skeleton-based models, with the latter being more robust with respect to the clip length.
This could result from the size of the feature vectors generated from the different encoders.
While VA-CNN produces feature vectors of size $2048$ for every clip, Swin only computes feature vectors of size $768$.
As the clip length increases, more information must be encoded in these feature vectors of constant size.
For the smaller sized feature vector of Swin, this might explain the bad performance for large clip lengths in PDAN compared to clips consisting of $16$ or less frames.

\subsection{Results for online usage}
\begin{figure}[t]
    \centering
    \vspace{2mm}
    \includegraphics[width=\hsize]{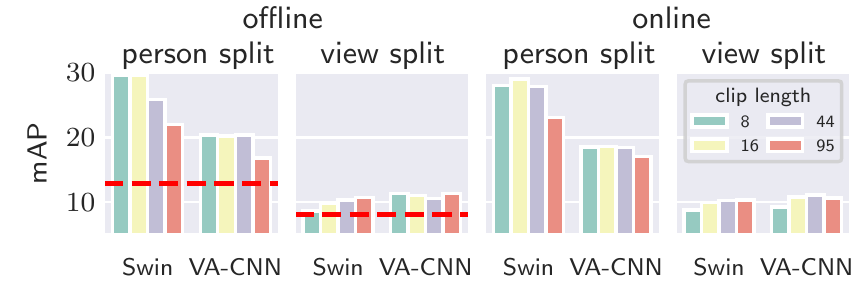}
    \vspace{-8mm}
    \caption{Results on the action detection task for offline and online usage achieved by PDAN with different backbones. The dashed red line marks results with the I3D encoder trained on Charades.
    The mAP results of the best models are reported in Tab.~\ref{tab:results_detection}
    }
    \vspace{-4mm}
    \label{fig:results_detection}
\end{figure}

For our robotic-application-specific evaluation we changed the PDAN model architecture so that the temporal receptive field for every frame only consists of present and past frames and no frames from the.
Following this change, we repeated our training and hyperparameter search and present our results in Fig.~\ref{fig:results_detection}.
As can be seen, the performance for online usage of PDAN is comparable to the offline usage and only marginally worse, which is to be expected, as the receptive field now views less relevant frames.

The clip size evaluation is important for the robotic application scenario, because the usage of PDAN with a higher clip size results in longer delay between updates on the current performed actions.
Fortunately, this evaluation shows that a smaller clip size, such as $16$ (ca. half a second at 30$\,$Hz), achieves very good results compared to the longer clip sizes, for video- and skeleton-based feature extractors.

\begin{table}[t]
    \centering
    \caption{Results on the action detection task for offline and online usage using PDAN with different encoders and modalities. We report frame-based mAP together with the clip length that achieved the best results.}
    \vspace{-2mm}
    \scriptsize
\begin{tabular}{lllrlrl}
\hline
                           &                          &                              & \multicolumn{2}{c}{offline usage}                         & \multicolumn{2}{c}{online usage}                               \\
                           & encoder                  & split                        & clip length                & \multicolumn{1}{r}{mAP}      & \multicolumn{1}{l}{clip length} & mAP                          \\ \hline
                           &                          & person                       & 16                         & 29.5                         & 16                              & 29.0                         \\
\multirow{-2}{*}{video}    & \multirow{-2}{*}{Swin}   & \cellcolor{lightgray!50!white!80}view & \cellcolor{lightgray!50!white!80}95 & \cellcolor{lightgray!50!white!80}10.7 & \cellcolor{lightgray!50!white!80}44      & \cellcolor{lightgray!50!white!80}10.3 \\ \hline
                           &                          & person                       & 8                          & 20.4                         & 16                              & 18.5                         \\
\multirow{-2}{*}{skeleton} & \multirow{-2}{*}{VA-CNN} & \cellcolor{lightgray!50!white!80}view & \cellcolor{lightgray!50!white!80}8 & \cellcolor{lightgray!50!white!80}11.3 & \cellcolor{lightgray!50!white!80}44      & \cellcolor{lightgray!50!white!80}11.1 \\ \hline
\end{tabular}
    \label{tab:results_detection}
    \vspace{-4mm}
\end{table}
\section{Conclusion}

In this paper, we presented the new ATTACH dataset containing actions performed during assembly.
In contrast to existing datasets, we labeled fine-grained actions for each hand individually, resulting in more than 68\% of overlapping annotations.
Based on our three camera setup, we defined a person and a view split which represent different challenges for action understanding models.

To create a first baseline, we reported results of state-of-the-art methods for both action recognition and detection on our state-of-the-art multi-label assembly dataset, using video and skeleton-sequence inputs respectively.
Furthermore, we also evaluated action detection for the cooperative robotic application task of achieving situational awareness for assistance of a worker during assembly.

\quad \textit{Future directions:}
The ATTACH dataset provides a lot of open potential regarding further action understanding tasks. 
E.g., the estimation of hand poses could provide further information for skeleton-based approaches when trying to perceive actions focused on the fingers, which are only very roughly represented by the Azure Kinect skeletons.
For this, our high resolution recordings serve as a good foundation.

\phantom{.}\\
To summarize, by providing the ATTACH dataset, we build a foundation for better action perception in the context of assembly tasks, which will contribute to the emerging field of human-robot collaboration.

\bibliographystyle{IEEEtran}
\bibliography{egbib}

\end{document}